%% file: main.tex
\documentclass[letterpaper, 10 pt, conference]{ieeeconf}
\IEEEoverridecommandlockouts 

\usepackage[utf8]{inputenc}
\usepackage[T1]{fontenc}
\usepackage{amsmath,amsfonts}
\usepackage{amssymb}
\usepackage{mathtools}
\usepackage{booktabs}
\usepackage{array}
\usepackage{multirow} 
\usepackage{makecell}
\usepackage{tabularx}
\usepackage{threeparttable}
\usepackage[table,xcdraw]{xcolor}

\usepackage{graphicx}
\usepackage{caption}
\usepackage{subcaption}
\usepackage{stfloats}
\usepackage{adjustbox}

\usepackage{algorithm}
\usepackage{algorithmicx}
\usepackage{algpseudocode}

\usepackage{textcomp}
\usepackage{url}
\usepackage{verbatim}
\usepackage{cite}

\definecolor{rblue}{rgb}{0,0.5,1}
\definecolor{awesome}{rgb}{1.0, 0.13, 0.32}
\definecolor{hollywoodcerise}{rgb}{0.96, 0.0, 0.63}
\definecolor{lasallegreen}{rgb}{0.03, 0.47, 0.19}
\definecolor{hanpurple}{rgb}{0.32, 0.09, 0.98}
\definecolor{green(pigment)}{rgb}{0.0, 0.65, 0.31}

\definecolor{bestbg}{HTML}{DDF7E3}
\definecolor{secondbg}{HTML}{FFE9E9}

\definecolor{grpName}{HTML}{F0F7FF}
\definecolor{grpMask}{HTML}{FFF7EC}
\makeatletter
\let\NAT@parse\undefined
\makeatother
\usepackage[pagebackref=false,breaklinks=true,colorlinks,bookmarks=false]{hyperref}
\hypersetup{colorlinks=true,
    linkcolor={red},
    citecolor={hanpurple},
    urlcolor={magenta}
}

\title{\LARGE \bf
Segment-to-Act: Label-Noise-Robust Action-Prompted Video Segmentation Towards Embodied Intelligence
}

\author{Wenxin Li$^{1}$, Kunyu Peng$^{2,3,*}$, Di Wen$^{2}$, Ruiping Liu$^{2}$, Mengfei Duan$^{1}$, Kai Luo$^{1}$, and Kailun Yang$^{1,*}$
\thanks{This work was supported in part by the National Natural Science Foundation of China (Grant No. 62473139), in part by the Hunan Provincial Research and Development Project (Grant No. 2025QK3019), in part by the State Key Laboratory of Autonomous Intelligent Unmanned Systems (the opening project number ZZKF2025-2-10), and in part by the Deutsche Forschungsgemeinschaft (DFG, German Research Foundation) - SFB 1574 - 471687386.}
\thanks{$^{1}$The authors are with the School of Artificial Intelligence and Robotics and the National Engineering Research Center of Robot Visual Perception and Control Technology, Hunan University, China (email: kailun.yang@hnu.edu.cn).}%
\thanks{$^{2}$The authors are with the Institute for Anthropomatics and Robotics, Karlsruhe Institute of Technology, Germany (email: firstname.lastname@kit.edu).}%
\thanks{$^{3}$The author is also with INSAIT, Sofia University ``St. Kliment Ohridski'', Bulgaria.}%
\thanks{*Corresponding authors: Kailun Yang and Kunyu Peng.}%
}

\begin{document}

\maketitle
\thispagestyle{empty}
\pagestyle{empty}

\begin{abstract}
Embodied intelligence relies on accurately segmenting objects actively involved in interactions. Action-based video object segmentation addresses this by linking segmentation with action semantics, but it depends on large-scale annotations and prompts that are costly, inconsistent, and prone to multimodal noise such as imprecise masks and referential ambiguity. To date, this challenge remains unexplored. In this work, we take the first step by studying action-based video object segmentation under label noise, focusing on two sources: textual prompt noise (category flips and within-category noun substitutions) and mask annotation noise (perturbed object boundaries to mimic imprecise supervision). Our contributions are threefold. First, we introduce two types of label noises for the action-based video object segmentation task. Second, we build up the first action-based video object segmentation under a label noise benchmark \textit{ActiSeg\mbox{-}NL} and adapt six label-noise learning strategies to this setting, and establish protocols for evaluating them under textual, boundary, and mixed noise. Third, we provide a comprehensive analysis linking noise types to failure modes and robustness gains, and we introduce a Parallel Mask Head Mechanism (\textit{PMHM}) to address mask annotation noise. Qualitative evaluations further reveal characteristic failure modes, including boundary leakage and mislocalization under boundary perturbations, as well as occasional identity substitutions under textual flips. Our comparative analysis reveals that different learning strategies exhibit distinct robustness profiles, governed by a foreground-background trade-off where some achieve balanced performance while others prioritize foreground accuracy at the cost of background precision. These results establish a clear sensitivity profile of action-based video object segmentation to imperfect annotations and set a benchmark for studying noise-robust learning in embodied perception. The established benchmark and source code will be made publicly available at \url{https://github.com/mylwx/ActiSeg-NL}.

\end{abstract}

\section{Introduction}
\input{Contents/Introduction}

\section{Related Work}
\input{Contents/Related_Work}

\section{Benchmark}
\input{Contents/Benchmark}
\section{Methodology}
\input{Contents/Methodology}

\section{Experiments}\label{sec:experiments}
\input{Contents/Experiments}

\section{Conclusion}
\input{Contents/Conclusion}

{\small
\bibliographystyle{IEEEtran}
\bibliography{bib}
}

\end{document}

%% file: Contents/Introduction.tex
\begin{figure}[!t]
\small
  \centering
  \includegraphics[width=\linewidth]{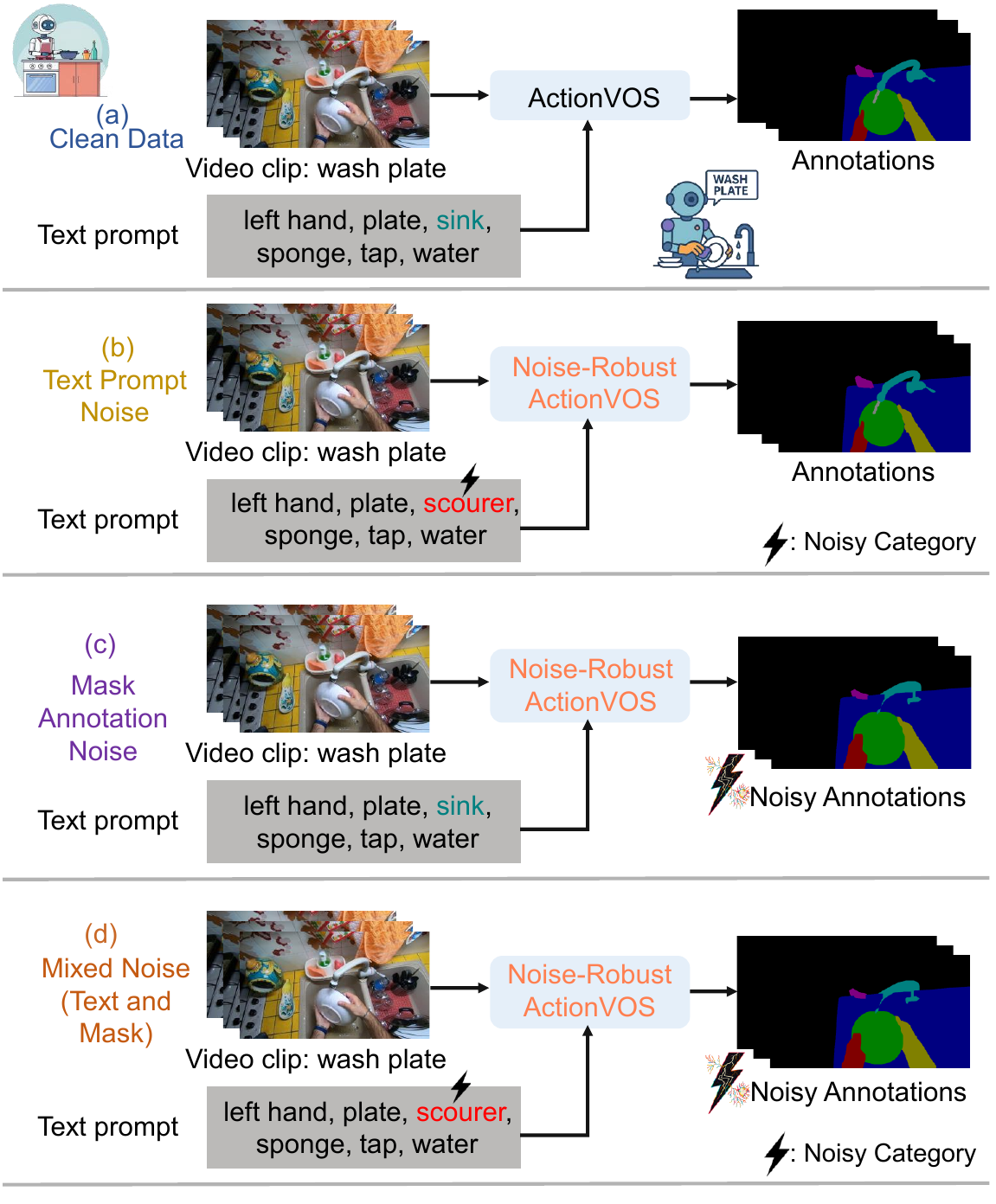}
  \caption{Training pipeline of action-based video object segmentation (ActionVOS) with one clean reference (a) and three controlled noise scenarios defined by \textit{ActiSeg\mbox{-}NL}, including text prompt noise (b), mask annotation noise (c), and their mixed condition (d). These scenarios approximate perception disturbances observed in egocentric recordings and in robot manipulation, and they test whether segmentation remains stable enough to support downstream action. We then apply and evaluate various adapted noise-robust strategies on each scenario for their mitigation effectiveness.}
  \label{fig:concept_noisy_actionvos}
  \vspace{-7mm}
\end{figure}

Video Object Segmentation (VOS) is a core perception substrate for embodied intelligence because it anchors language instructions to pixel-level object representations that support planning and control in instruction following, rearrangement, and manipulation settings~\cite{grauman2022ego4d}. VOS has progressed from classical semi-supervised settings in which the target object is specified by a mask in the first frame to language-guided paradigms~\cite{Caelles_2017_CVPR,Oh_2019_ICCV}. 
Referring Video Object Segmentation extends VOS by conditioning object segmentation on free-form textual expressions that describe specific targets in a video~\cite{botach2022end,wu2022language}. As a specialization of R-VOS, action-based video object segmentation (ActionVOS)~\cite{ouyang2024actionvos} refines the task by employing structured action narrations as prompts to segment objects actively engaged in human actions. This design facilitates downstream applications in human-robot interaction and embodied perception, where agents must select task-relevant objects and execute manipulation under egocentric conditions~\cite{yenamandra2023homerobot}.

However, deploying action-based video object segmentation in realistic scenarios presents significant practical challenges. Specifically, large-scale mask annotation in egocentric videos is prohibitively expensive~\cite{chen2024masktrack}, with annotation effort growing proportionally to object density and video length~\cite{damen2022rescaling}. Additionally, annotations often suffer from subjective inconsistencies across annotators, further complicating data reliability~\cite{ye2024learning}. Meanwhile, linguistic prompts also introduce referential noise, including lexical confusions such as \textit{``flour''} versus \textit{``four''} observed in VISOR~\cite{darkhalil2022epic}. Despite its practical importance, such multimodal noise remains largely underexplored, limiting the robustness and generalizability of current action-based video object segmentation models.

Existing noisy-label learning methods have primarily focused on image classification~\cite{sheng2024adaptive,liu2020early}, with limited extensions to video segmentation tasks~\cite{cho2025spatial}. Moreover, conventional strategies such as sample selection~\cite{miao2024learning} and robust loss functions~\cite{zhang2018generalized,zhou2023asymmetric} are ill-suited for action-based video object segmentation, as they fail to simultaneously handle the intertwined pixel-level and language-level label noise. Linguistic noise, in particular, introduces semantic ambiguities and referential mismatches~\cite{yan2024referred}, posing unique challenges beyond pixel-level errors.  

In order to fill this gap, we introduce \textit{ActiSeg\mbox{-}NL}, a benchmark designed explicitly to assess the robustness of action-based video object segmentation. 
Fig.~\ref{fig:concept_noisy_actionvos} depicts the action-based video object segmentation training pipeline. \textit{ActiSeg\mbox{-}NL} injects synthetic noise into the VISOR training split and defines three controlled scenarios, namely text prompt noise, mask annotation noise, and a mixed condition. We adapt a wide spectrum of noise-robust learning strategies to the action-based video object segmentation setting, including Co-teaching~\cite{han2018co}, Generalized Cross Entropy (GCE)~\cite{zhang2018generalized}, Symmetric Cross Entropy (SCE)~\cite{wang2019symmetric}, Active Passive Loss (APL)~\cite{ye2023active}, Early Learning Regularization (ELR)\cite{liu2020early}, and NPN, which integrates partial label learning and negative learning~\cite{sheng2024adaptive}.
Our benchmark reveals the limitations of these methods under text prompt noise, mask annotation noise, and the mixed condition within action-based video object segmentation, and underscores the need for solutions tailored to embodied segmentation.

Our contributions are threefold:
\begin{itemize}

\item We initiate the study of noisy labels in action-based video object segmentation by formalizing \emph{text prompt noise}, \emph{mask annotation noise}, and \emph{mixed noise}, establishing a noisy-label taxonomy that links semantic ambiguity in language prompts with boundary imprecision in pixel masks.

\item We introduce \textit{ActiSeg\mbox{-}NL}, the first benchmark for this setting, and unify adaptations of a broad range of diverse noisy-label learning strategies, including Co-teaching~\cite{han2018co}, GCE~\cite{zhang2018generalized}, SCE~\cite{wang2019symmetric}, APL~\cite{ye2023active}, ELR~\cite{liu2020early}, and NPN~\cite{sheng2024adaptive}, to the pixel-level and language-conditioned setting, providing the first large-scale comparison of noise-robust training in action-based video object segmentation. 

\item Our benchmark analysis shows that robustness is not monolithic. Text noise challenges grounding and reduces foreground coverage, while Co-teaching preserves foreground regions, whereas mask noise degrades overlap, and APL improves overlap. Under mixed conditions, pixelwise losses (\textit{e.g.}, GCE, SCE) are more stable than sample filtering, with ELR increasing positive regions at the cost of background precision. Additionally, we propose \textit{Parallel Mask Head Mechanism} (\textit{PMHM}), a parallel-head consistency scheme that mitigates mask noise in action-based video object segmentation.  

\end{itemize}

%% file: Contents/Related_Work.tex
\subsection{Referring Video Object Segmentation}
Referring Video Object Segmentation (R-VOS)~\cite{botach2022end,wu2022language} aims to accurately segment target objects in video sequences using text prompts, enabling applications in human-computer interaction~\cite{joo2024hand} and video analysis~\cite{liang2025referdino}. 
Recent frameworks have considerably pushed forward progress in this domain.
MTTR~\cite{botach2022end} introduces an end-to-end multimodal Transformer that simplifies previously multi-stage workflows. ReferFormer~\cite{wu2022language} employs language prompts as dynamic queries for efficient mask generation. For handling complex scenes, SgMg~\cite{miao2023spectrum} incorporates spectrum-guided cross-modal fusion.
FindTrack~\cite{cho2025find} enhances robustness by explicitly decoupling object recognition from mask propagation.
Action-based video object segmentation extends R-VOS by incorporating action narrations into text prompts to segment only objects that actively participate in the ongoing action (\textit{e.g.}, narrated target objects, hands, handheld tools, and involved containers and contents), reducing mis-segmentation of inactive objects~\cite{ouyang2024actionvos}.
While existing datasets lack explicit participation labels and thus depend on potentially erroneous pseudo labels, Action-based video object segmentation mitigates noise in active versus inactive labels~\cite{ouyang2024actionvos}, yet robustness to text prompt noise and mask annotation noise remains underexplored.
We introduce \textit{ActiSeg\mbox{-}NL}, the first benchmark that systematically considers action-based video object segmentation under text prompt noise, mask annotation noise, and their mixed setting.

\subsection{Learning with Noisy Labels}
Noisy labels~\cite{song2022learning,liu2022adaptive}, prevalent in large-scale datasets, significantly impair deep neural network performance due to their reliance on high-quality annotations. 
Song~\textit{et al.}~\cite{song2022learning} study symmetric and asymmetric noise in image classification.
Luo~\textit{et al.}~\cite{luo2022deep} leverage high-level spatial structures as supervisory signals to mitigate mislabeled annotations in segmentation tasks. 
Feng~\textit{et al.}~\cite{feng2023weakly} enhance weakly supervised segmentation with online pseudo-mask correction. 
In video object segmentation, Enki~\textit{et al.}~\cite{cho2025spatial} propose a space-mask-based framework to correct noise using spatial context, whereas Kimhi~\textit{et al.}~\cite{kimhi2025noisyannotationssemanticsegmentation} establish benchmarks like COCO-N and VIPER-N~\cite{kimhi2025noisyannotationssemanticsegmentation} to evaluate instance segmentation robustness against class confusion and boundary distortion. 
However, many of these methods~\cite{song2022learning,liu2022adaptive} operate at the image level, and most pixel-level approaches~\cite{yao2023learning,feng2023weakly,cho2025spatial} do not incorporate text guidance, so they are not directly applicable to action-based video object segmentation, which requires supervision at the pixel level conditioned on language and vision. We fill this gap by formalizing label noise for text prompts and mask annotations, as illustrated in Fig.~\ref{fig:concept_noisy_actionvos}, and by introducing \textit{ActiSeg\mbox{-}NL}, the first benchmark that unifies adaptations of diverse noisy-label learning strategies for action-based video object segmentation.

%% file: Contents/Benchmark.tex
\label{sec:benchmark}
\textit{ActiSeg\mbox{-}NL} is a benchmark for evaluating robustness in action-based video object segmentation under text, mask, and mixed noise, focusing on failure modes in embodied perception. We generate these controlled noises on the VISOR~\cite{darkhalil2022epic} training split for model training and evaluate on the original clean data. This section details our noise construction methods.
\begin{algorithm}[t]
\small
\caption{Text Prompt Noise Generation}
\label{alg:text_prompt_noise}
\textbf{Input:} Clean Category $\mathcal{C}$, noise rate $\rho$, class mapping $\mathcal{M}: \text{class} \to \{\mathcal{C}_1, \mathcal{C}_2, \dots\}$, classes $\Omega = \{1, \dots, K\}$. \\
\textbf{Output:} Noisy category $\mathcal{C}_{\text{noise}}$. \\
\textbf{Initialize:} $\text{class}_{\mathcal{C}} \gets \text{class of } \mathcal{C}$.

\begin{algorithmic}[1]
\If{$\text{Uniform}(0,1) < \rho$} \Comment{Class flipping}
    \State Select $\text{class}_{\text{flip}} \in \Omega \setminus \{\text{class}_{\mathcal{C}}\}$ uniformly at random.
    \State Set $\mathcal{C}_{\text{noise}} \gets \text{random category from } \mathcal{M}(\text{class}_{\text{flip}})$.
\Else \Comment{Synonymous replacement}
    \State Set $\mathcal{C}_{\text{noise}} \gets \text{random category from } \mathcal{M}(\text{class}_{\mathcal{C}})$.
\EndIf
\State \textbf{Return:} $\mathcal{C}_{\text{noise}}$.
\end{algorithmic}
\end{algorithm}

\begin{figure}[tb]
\small 
  \centering
  \includegraphics[width=\linewidth]{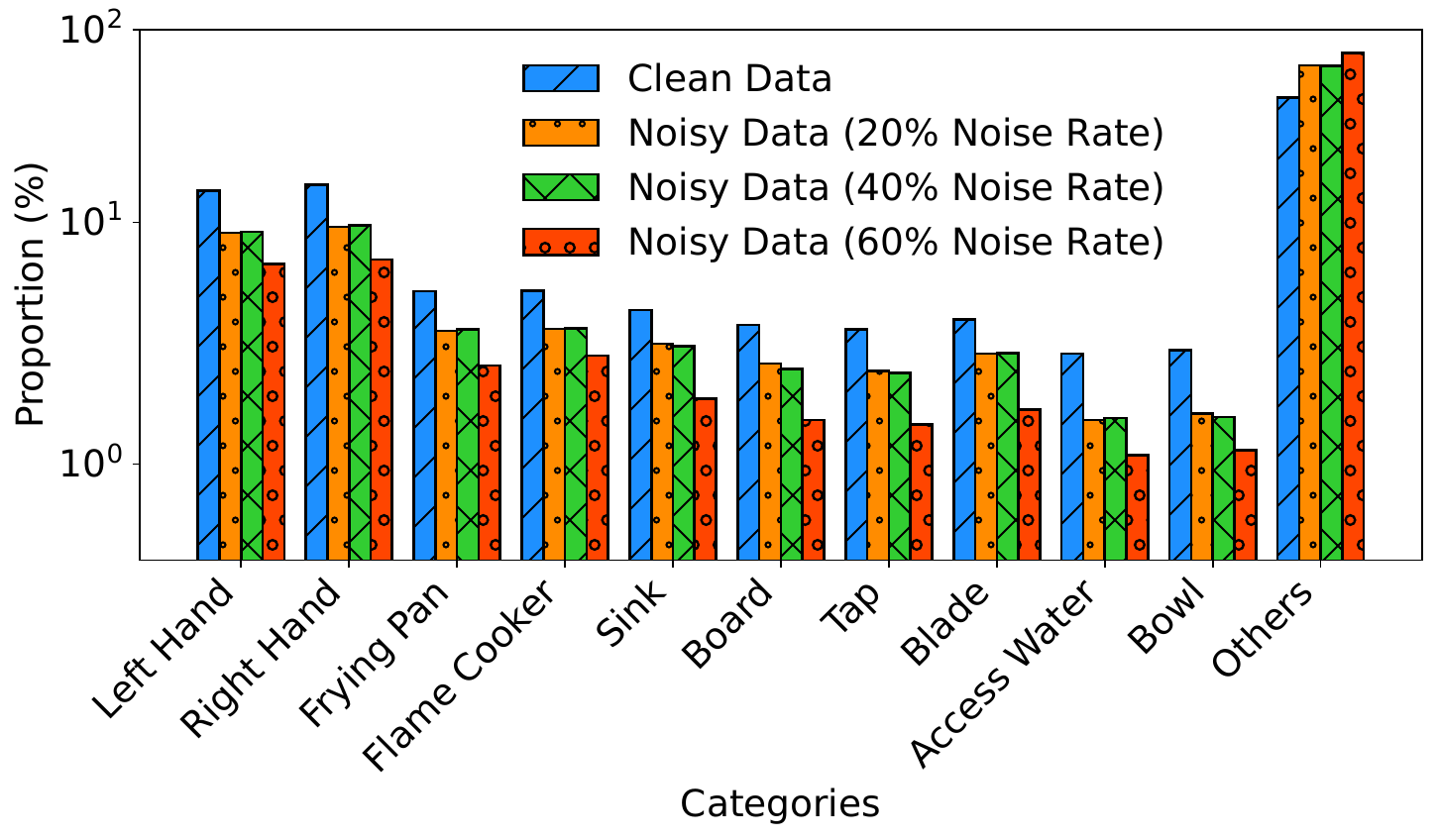}
  \caption{Comparison of category distributions before and after applying text prompt noise at $20\%$, $40\%$, and $60\%$ noise rates for major categories (proportion ${>}1\%$) and Others.}
  \label{fig:name_category}
  \vspace{-5mm}
\end{figure}
\subsection{Text Prompt Noise}

\noindent \textbf{Construction.}
We generate text prompt noise for the VISOR training split on target objects with two components. First, symmetric noise is built by randomly flipping referent objects in text prompts, such as replacing \textit{``container''} with \textit{``fridge''} in the prompt \textit{``container used in the action of take celery''} with probabilities of $20\%$, $40\%$, or $60\%$, which represent low, medium, and high noise levels commonly used in noisy label learning to assess robustness, following \cite{ye2023active, liu2020early}. 
The target category is randomly selected from a predefined set to ensure equal replacement probability. 
Second, we substitute nouns with synonymous or related terms within the same category, such as replacing \textit{``container''} with \textit{``food container''} or \textit{``cheese container''} to increase semantic complexity, by uniformly sampling the replacement from $\mathcal{M}(\text{class}_{\mathcal{C}})$ when the class flipping condition is not met, as described in Alg.~\ref{alg:text_prompt_noise}. The whole procedure is summarized in Alg.~\ref{alg:text_prompt_noise}.

\noindent \textbf{Analysis.}
We diagnose distribution shifts induced by text prompt noise. To validate this noise strategy, we analyze category distributions in the original and noisy training sets. Fig.~\ref{fig:name_category} compares major categories (proportion ${>}1\%$ in the original set) and a summarized \textit{``Others''} category across clean data and $20\%$, $40\%$, and $60\%$ noise rates. 
These statistics show that \textit{``left hand''} and \textit{``right hand''} proportions decrease from $14.58\%$ and $15.60\%$ to $6.08\%$ and $6.39\%$ at  the $60\%$ noise rate, respectively, while \textit{``Others''} increases from $44.43\%$ to $75.47\%$. 
This shift validates the disruptive effect of our noise strategy, as depicted in Fig.~\ref{fig:concept_noisy_actionvos}, establishing \textit{ActiSeg\mbox{-}NL} as a dedicated benchmark for exploring text prompt noise in video segmentation.

\begin{figure}[tb]
\small 
  \centering
  \includegraphics[width=\linewidth]{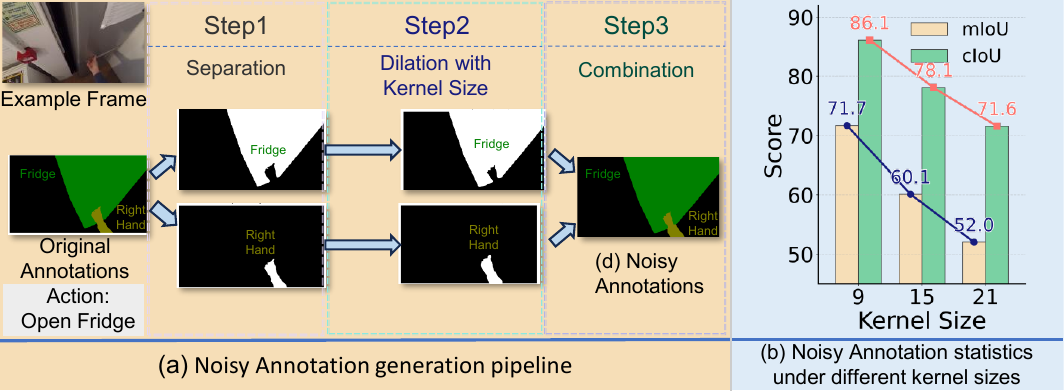}
  \caption{Mask annotation noise generation and severity statistics in \textit{ActiSeg\mbox{-}NL}. (a) Generation pipeline: clean instance masks are separated, dilated with different kernel sizes, and recombined to produce noisy annotations. (b) Summary statistics: larger kernel leads to lower mIoU and cIoU, quantifying the decline in annotation quality.}
  \label{fig:mask_noise}
  \vspace{-4mm}
\end{figure}

\subsection{Mask Annotation Noise}
\noindent \textbf{Construction.}
We employ a separate-dilate-combine strategy to simulate mask annotation noise, replicating realistic boundary errors in crowd-sourced annotations for positive objects in mask segmentation tasks. First, we separate clean semantic segmentation annotations into binary masks for each object, such as \textit{``right hand''} and \textit{``fridge''} in the \textit{``open fridge''} action. Next, we apply morphological dilation with a square kernel (kernel sizes $9$, $15$, or $21$) to expand object boundaries, mimicking the boundary blur observed in human annotations. 
Finally, we combine the dilated per-object masks into noisy annotations using a first-hit rule with a fixed object order, assigning each pixel the label of the first dilated mask that covers it while disregarding subsequent overlaps, thereby ensuring a single label per pixel.
The left of Fig.~\ref{fig:mask_noise} illustrates this process with an example frame. 
The entire procedure is summarized in Alg.~\ref{alg:mask_noise_generation}.

\begin{figure*}[tb]
\small 
  \centering
  \includegraphics[width=\textwidth]{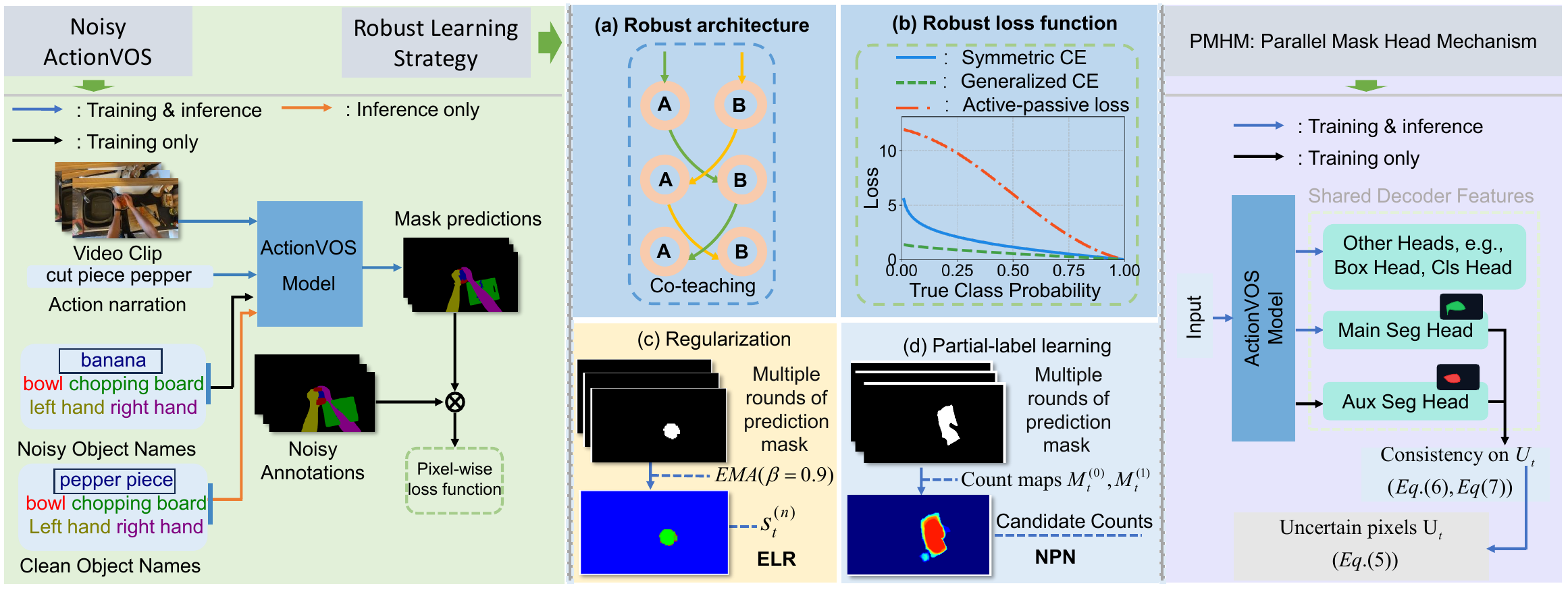}
  \caption{Overview of the noisy action-based video object segmentation framework and robustness strategies. \textbf{Left:} the framework consumes video frames, noisy object names (\textit{e.g.}, ``banana''), and action narrations (\textit{e.g.}, ``cut piece pepper''), to predict a pixel-level mask. 
  \textbf{Middle:} four complementary strategies, (a) \emph{Co-teaching}, where two networks exchange small-loss samples to suppress label noise, (b) \emph{Noise-robust Losses} GCE, SCE, and APL that balance accuracy and robustness, (c) \emph{ELR}, which mitigates overfitting to noisy annotations with an EMA-based regularizer, (d) \emph{NPN}, which integrates candidate-set reasoning with negative learning and with consistency across weak and strong views for pixel supervision. \textbf{Right:} \textit{PMHM} architecture. During training, a lightweight auxiliary head runs in parallel with the main head to achieve prediction consistency on uncertain pixels, using symmetric KL divergence across heads and decoder layers.}
  \label{fig:noisy_actionvos_flow}
  \vspace{-6mm}
\end{figure*}

\begin{algorithm}[t]
\small
\caption{Mask Annotation Noise Generation}
\label{alg:mask_noise_generation}
\textbf{Input:} Multi-label annotation mask $\mathcal{A}$, positive object IDs $\Omega = \{ \text{obj}_1, \text{obj}_2, \dots, \text{obj}_N \}$, kernel size $k \in \{9, 15, 21\}$. \\
\textbf{Output:} Noisy annotation mask $\mathcal{A}_{\text{noise}}$. \\
\textbf{Initialize:} Empty list $\mathcal{R} \leftarrow \emptyset$, $\mathcal{A}_{\text{noise}} \leftarrow \mathbf{0}^{H \times W}$ (zero-initialized mask of same size as $\mathcal{A}$).

\begin{algorithmic}[1]
\State \textbf{Separate:} For each $\text{obj}_i \in \Omega$, compute binary mask $\mathcal{B}_i$ where $\mathcal{B}_i[x, y] = 1$ if $\mathcal{A}[x, y] = \text{obj}_i$, else $\mathcal{B}_i[x, y] = 0$.
\For{each $\text{obj}_i \in \Omega$}
    \State Compute binary mask $\mathcal{B}_i$ from $\mathcal{A}$ for $\text{obj}_i$.
    \State Apply morphological dilation to $\mathcal{B}_i$ with square kernel of size $k$, yielding dilated mask $\mathcal{D}_i$.
    \State Append $\mathcal{D}_i$ to $\mathcal{R}$.
\EndFor
\State \textbf{Combine:} Initialize $\mathcal{A}_{\text{noise}}$ as a zero mask of size $H \times W$.
\For{each position $[x, y]$ in $\mathcal{A}_{\text{noise}}$}
    \For{each dilated mask $\mathcal{D}_i \in \mathcal{R}$}
        \If{$\mathcal{A}_{\text{noise}}[x, y] = 0$ and $\mathcal{D}_i[x, y] \neq 0$}
            \State Set $\mathcal{A}_{\text{noise}}[x, y] \leftarrow \text{obj}_i$.
        \EndIf
    \EndFor
\EndFor
\State \textbf{Return:} $\mathcal{A}_{\text{noise}}$.
\end{algorithmic}
\vspace{-2pt}
\end{algorithm}

\noindent \textbf{Analysis.}
By merging the dilated per-object masks with a first hit rule under a fixed object order, overlaps create boundary competition that resembles the blur observed in crowdsourced annotations, yet the resulting noise is more challenging than typical crowdsourced boundary errors and grows stronger as the dilation kernel increases. 
The severity of the generated mask noise is directly controlled by the dilation kernel size. Increasing the kernel size from $9$ to $21$ monotonically degrades annotation quality (Fig.~\ref{fig:mask_noise}, right). 

%% file: Contents/Methodology.tex
\noindent\textbf{Setup and Notation.}
We study action-based video object segmentation~\cite{ouyang2024actionvos} with noisy supervision. The task is to generate a per-frame binary mask identifying the active object, given a video clip and a textual prompt. Formally, for each pixel \((x,y)\) in a frame \(t\), the model predicts a foreground probability \(p_t(x,y)\in[0,1]\). This prediction is trained against a corresponding ground truth label \(y_t(x,y)\in\{0,1\}\), which may be noisy. Training minimizes a frame-averaged pixel-wise objective to ensure reliable localization for embodied manipulation.

\subsection{Adapted Robust Learners for Pixel-level and Language-conditioned Segmentation}\label{sec:adopted}
We adapt and integrate a broad set of noisy label strategies into the action-based video object segmentation setting~\cite{ouyang2024actionvos}. 
Each method is independently adapted to operate on pixel probabilities and language-conditioned features. For all methods, the training objective averages per pixel losses over $\mathcal{I}_t$ and over time. 
The adapted strategies include Co-teaching~\cite{han2018co}, GCE~\cite{zhang2018generalized}, SCE~\cite{wang2019symmetric}, APL~\cite{ye2023active}, ELR~\cite{liu2020early}, and NPN~\cite{sheng2024adaptive}, enabling the first large-scale comparison of noise-robust training in action-based video object segmentation~\cite{ouyang2024actionvos}.

\textit{Co-teaching}~\cite{han2018co} suppresses heavy prompt mismatch by peer selection of small loss samples between two networks. 
At epoch \(n\), each network selects a keep set containing the lowest loss samples in its mini-batch, and its size follows the conservative keep rate \(R_n = 1 - \rho\,\min\!\bigl(n/T_k,\,1\bigr)\). 
The selected samples are then used to supervise the peer network.

GCE~\cite{zhang2018generalized} reduces the impact of outlying pixels by interpolating between cross entropy and mean absolute error. Define the correctness score $r=y_t p_t+(1-y_t)(1-p_t)$ and use $\ell_{\mathrm{GCE}}=(1-r^{\,q})/q$ with $q\in(0,1]$.

SCE~\cite{wang2019symmetric} balances the standard cross entropy \textit{loss} $\mathcal{L}_{\mathrm{CE}}=\frac{1}{\sum_t |\mathcal{I}_t|}\sum_{t}\sum_{(x,y)\in \mathcal{I}_t}\ell_{\mathrm{CE}}^{\,txy}$ and the reverse cross entropy \textit{loss} $\mathcal{L}_{\mathrm{RCE}}=\frac{1}{\sum_t |\mathcal{I}_t|}\sum_{t}\sum_{(x,y)\in \mathcal{I}_t}\ell_{\mathrm{RCE}}^{\,txy}$ for stability when foreground and background are partially corrupted, with weights $\alpha$ and $\beta$ in the combined objective $\mathcal{L}_{\text{sym}}=\alpha\,\mathcal{L}_{\mathrm{CE}}+\beta\,\mathcal{L}_{\mathrm{RCE}}$.

APL~\cite{ye2023active} strengthens confident decisions and passively penalizes ambiguous ones. We use the active $\frac{1-r^{\,q}}{q}$ and the passive $|p_t-y_t|$ in a weighted sum.

\input{table/name_prompt_noise}
\textit{ELR}~\cite{liu2020early} prevents noisy label memorization by regularizing its predictions toward their historical average. It combines a primary classification loss with a temporal agreement regularizer. The primary loss is the focal loss~\cite{lin2017focal}, which uses weights $w^+{=}\alpha(1{-}p_t^{(n)})^\gamma$ and $w^-{=}(1{-}\alpha)(p_t^{(n)})^\gamma$ to down-weight well-classified examples:
\begin{equation}\label{eq:elr-focal}
\ell_{\text{focal}}^{\,txy}=-\bigl[y_t\,w^+ \log p_t^{(n)} + (1-y_t)\,w^- \log\bigl(1-p_t^{(n)}\bigr)\bigr].
\end{equation}
To enforce consistency, an exponential moving average (EMA) of predictions, $s_t^{(n)}$, is maintained across training epochs $n$, updated via
$s_t^{(n)}=(1-\beta)\,s_t^{(n-1)}+\beta\,p_t^{(n)}$.
An agreement score $d_t^{(n)}=p_t^{(n)} s_t^{(n)}+\bigl(1-p_t^{(n)}\bigr)\bigl(1-s_t^{(n)}\bigr)$ measures the similarity between the current prediction and its historical EMA.
This score is used to formulate a regularization term $R^{txy}=-\log\bigl(1-d_t^{(n)}+\epsilon\bigr)$, where $\epsilon$ ensures numerical stability. The final objective is a weighted sum of the focal loss and this regularizer, $\ell_{\text{focal}}^{\,txy}+\lambda R^{txy}$, averaged over all pixels and frames.

\textit{NPN}~\cite{sheng2024adaptive} integrates partial label learning, negative learning, and dual-view consistency at the pixel level. For each pixel, we form candidate indicators $A_{t}^{(1,n)}=y_t+\mathbb{I}\!\bigl[p_t^{(n)}>0.5\bigr]$ and $A_{t}^{(0,n)}=(1-y_t)+\mathbb{I}\!\bigl[p_t^{(n)}\le 0.5\bigr]$ and accumulate counts $M_t^{(1)}\leftarrow M_t^{(1)}+A_{t}^{(1,n)}$ and $M_t^{(0)}\leftarrow M_t^{(0)}+A_{t}^{(0,n)}$. The proxy label is $y_t^\star=\mathbb{I}[M_t^{(1)}\ge M_t^{(0)}]$ and the reliability is $w_t=\dfrac{\max(M_t^{(1)},M_t^{(0)})}{M_t^{(1)}+M_t^{(0)}}$. The partial label loss $\ell_{\text{PLL}}^{\,txy}$ is
\begin{equation}\label{eq:npn-pll}
\ell_{\text{PLL}}^{\,txy}=-w_t\bigl(y_t^\star\log p_t^{(n)}+(1-y_t^\star)\log\bigl(1-p_t^{(n)}\bigr)\bigr),
\end{equation}
and the negative learning loss $\ell_{\text{NL}}^{\,txy}$ is
\begin{equation}\label{eq:npn-nl}
\begin{aligned}
\ell_{\text{NL}}^{\,txy}=&-\bigl(1-\mathbb{I}[A_{t}^{(1,n)}\ge 1]\bigr)\log\bigl(1-p_t^{(n)}\bigr) \\
&-\bigl(1-\mathbb{I}[A_{t}^{(0,n)}\ge 1]\bigr)\log p_t^{(n)}.
\end{aligned}
\end{equation}
Dual-view consistency $\ell_{\text{CR}}^{\,txy}$ applies weak-to-strong cross-entropy with hard targets from the weak view. The final loss averages $\ell_{\text{PLL}}^{\,txy}+\alpha\,\ell_{\text{NL}}^{\,txy}+\beta\,\ell_{\text{CR}}^{\,txy}$ over pixels and frames.
\input{table/mask_edge_noise}

\subsection{Proposed PMHM Consistency with Main and Auxiliary Mask Heads}\label{sec:pmhm}
We propose the Parallel Mask Head Mechanism (\textit{PMHM}), a memory-efficient module for noisy label segmentation that avoids dual models~\cite{han2018co} and storing historical predictions for each pixel~\cite{lin2017focal,sheng2024adaptive}, both of which incur a high GPU memory burden. It features a main head and a parallel, lightweight
auxiliary head that share decoder features. 

During training, we apply mild perturbations (random dropout and decaying-rate freezing) to the auxiliary head to foster prediction diversity. The auxiliary head is discarded at inference, incurring no computational overhead.
Our method identifies uncertain pixels $U_t$ as those near the decision boundary or with high spatial gradients:
\begingroup
\setlength{\jot}{1pt}%
\setlength{\abovedisplayskip}{6pt}%
\setlength{\belowdisplayskip}{6pt}%
\setlength{\abovedisplayshortskip}{4pt}%
\setlength{\belowdisplayshortskip}{4pt}%
\begin{equation}
\begin{aligned}
U_t=\{(x,y)\in\mathcal{I}_t \mid &\, |p_t^{\mathrm{m}}(x,y)-0.5|<\tau_m \\
& \ \ \vee \ \ \|\nabla p_t^{\mathrm{m}}(x,y)\|_2>\tau_e\}.
\end{aligned}
\end{equation}
\endgroup

The head consistency loss, $L_{\text{head}}$, aligns the predictions of the main ($p_t^{\mathrm{m}}(x,y)$)  and auxiliary ($p_t^{\mathrm{a}}(x,y)$) heads on these uncertain pixels using a symmetric KL divergence:
\begingroup
\setlength{\jot}{1pt}%
\setlength{\abovedisplayskip}{6pt}%
\setlength{\belowdisplayskip}{6pt}%
\setlength{\abovedisplayshortskip}{4pt}%
\setlength{\belowdisplayshortskip}{4pt}%
\begin{equation}
\begin{aligned}
L_{\text{head}}=\frac{1}{\sum_t |U_t|}\sum_t \sum_{(x,y)\in U_t} \Bigl[&
D_{\mathrm{KL}}\!\big(p_t^{\mathrm{m}}\|p_t^{\mathrm{a}}\big) \\
&+ D_{\mathrm{KL}}\!\big(p_t^{\mathrm{a}}\|p_t^{\mathrm{m}}\big)\Bigr].
\end{aligned}
\end{equation}
\endgroup
We also introduce a layer consistency loss, $L_{\text{layer}}$, to align predictions from early decoder stages ($p_t^{\mathrm{m},\ell}$) with the final stage prediction ($p_t^{\mathrm{m},L}$) on the same uncertain pixels $U_t$:
\begin{equation}
\begin{aligned}
L_{\text{layer}} &=
\frac{1}{\sum_t |U_t|}
\sum_{t}\,\sum_{(x,y)\in U_t}\,
\frac{1}{L-1}\sum_{\ell=1}^{L-1}\Bigl[
D_{\mathrm{KL}}\bigl(p_t^{\mathrm{m},L}\,\|\,p_t^{\mathrm{m},\ell}\bigr) \\
&\qquad\qquad\qquad
+ D_{\mathrm{KL}}\bigl(p_t^{\mathrm{m},\ell}\,\|\,p_t^{\mathrm{m},L}\bigr)
\Bigr].
\end{aligned}
\end{equation}
The loss $L_{\text{seg}}$ combines a hard mask loss $L_{\text{hard}}$~\cite{ouyang2024actionvos} on confident pixels (outside $U_t$) with our consistency terms:
\begin{equation}
L_{\text{seg}} = L_{\text{hard}} + \lambda_{\text{head}} L_{\text{head}} + \lambda_{\text{layer}} L_{\text{layer}}.
\end{equation}
Hyperparameters $\tau_m$, $\tau_e$, $\lambda_{\text{head}}$, and $\lambda_{\text{layer}}$ govern uncertainty selection and regularization: $\tau_m$ defines a margin near $0.5$ for low confidence pixels, $\tau_e$ thresholds $\|\nabla p_t^{\mathrm{m}}\|$ to capture boundaries, and $\lambda_{\text{head}}$ with $\lambda_{\text{layer}}$ weight head and layer consistency against $L_{\text{hard}}$; concrete values are given in Sec.~\ref{sec:experiments}.

%% file: table/name_prompt_noise.tex
\begin{table}[t]
\centering
\setlength{\tabcolsep}{3pt}  
\caption{Action-based video object segmentation model under text prompt noise (noise-rate sweep).}
\vspace{-4pt}
\label{tab:name_prompt_noise}
\begin{tabular}{r|cccccc}
\toprule[1pt]
\makecell{Noise Rate\\(\%)} & p-mIoU$\uparrow$ & n-mIoU$\downarrow$ & p-cIoU$\uparrow$ & n-cIoU$\downarrow$ & gIoU$\uparrow$ & Acc$\uparrow$ \\
\midrule[1pt]
0  & 65.0 & 19.7 & 71.8 & 34.1 & 70.7 & 81.7 \\
20 & 57.2 & 13.3 & 64.4 & 16.6 & 69.6 & 79.3 \\
40 & 51.2 & 11.0 & 57.9 & 14.2 & 67.6 & 75.3 \\
60 & 49.0 & 10.0 & 57.4 & 12.7 & 67.1 & 73.1 \\
\bottomrule[1pt]
\end{tabular}
\vspace{-16pt}
\end{table}

%% file: table/mask_edge_noise.tex
\begin{table}[ht]
\caption{Action-based video object segmentation model with noisy annotations.}
\vspace{-10pt}
\label{tab:mask_edge_noise}
\begin{center}
\setlength{\tabcolsep}{3pt}  
\begin{tabular}{r|cccccc}
\toprule [1pt]
kernel size  & p-mIoU$\uparrow$ & n-mIoU$\downarrow$ & p-cIoU$\uparrow$ & n-cIoU$\downarrow$ & gIoU$\uparrow$ & Acc$\uparrow$ \\ \midrule [1pt]
0 &  65.0 & 19.7 & 71.8 & 34.1 & 70.7 & 81.7 \\
$9{\times}9$  & 55.7 & 13.7 & 64.8 & 20.3 & 68.2 & 80.4 \\
$15{\times}15$  & 49.1 & 13.3 & 61.1 & 19.7 & 63.7 & 80.0 \\
$21{\times}21$  & 44.8 & 13.9 & 57.4 & 29.9 & 60.3 & 79.2 \\ \bottomrule [1pt]
\end{tabular}
\end{center}
\vspace{-24pt}
\end{table}

%% file: Contents/Experiments.tex
\subsection{Datasets and Metrics}
\textit{Datasets.} 
We train on the VISOR training split ($13{,}205$ clips; $76{,}873$ objects) with our synthetic noise (Section~\ref{sec:benchmark}), and evaluate on the clean action-based video object segmentation benchmark ($294$ manually annotated clips) following the protocol in~\cite{ouyang2024actionvos}. Details of dataset construction and noise simulation are provided in Section~\ref{sec:benchmark}.
\textit{Metrics.} 
Following~\cite{ouyang2024actionvos}, we use several metrics to evaluate performance. We report Mean IoU (mIoU) and cumulative IoU (cIoU) computed separately for active foregrounds (p-mIoU, p-cIoU) and inactive backgrounds (n-mIoU, n-cIoU). Additionally, we use generalized IoU (gIoU)~\cite{liu2023gres} for overall segmentation quality and Accuracy (Acc) to evaluate the classification of active versus inactive objects.

\input{table/noise_all_single}

\subsection{Implementation Details}
We use ResNet-101~\cite{he2016deep} as the visual encoder and RoBERTa~\cite{liu2019roberta} as the text encoder (frozen). 
All models are initialized from Ref-YouTube-VOS checkpoints~\cite{seo2020urvos}. The optimizer and loss coefficients follow~\cite{ouyang2024actionvos}.
Data augmentations follow~\cite{wang2021end}; for ELR~\cite{liu2020early} and NPN~\cite{sheng2024adaptive}, we disable spatial transforms and keep only pixel-wise photometric distortion and normalization to preserve per-pixel alignment required by historical prediction updates. Method-specific settings are: Co-teaching keeps samples with a loss rank ${\le}0.95$, uses gradient accumulation of $128{\times}$, and sets $T_k{=}6$; 
GCE and APL use $q{=}0.7$~\cite{zhang2018generalized}; ELR and NPN both use focal BCE and serialize per-pixel histories to disk, with ELR adding temporal regularization \((\beta{=}0.9,\ \epsilon{=}10^{-6},\ \alpha{=}0.5,\ \gamma{=}2,\ \lambda{=}1)\) and NPN applying thresholding \((\alpha{=}0.1,\ \beta{=}0.2)\). 
PMHM uses thresholds $\tau_m{=}0.20,\ \tau_e{=}0.85$; loss weights are $\lambda_{\text{head}}{=}0.1,\ \lambda_{\text{layer}}{=}0.1$. 
Training runs on $4{\times}$ RTX 3090 for $6$ epochs, except that Co-teaching uses $10$ epochs to ensure convergence of its progressive sample selection mechanism under noisy labels. 

\subsection{Influence of Text Prompt Noise}
Under a name-perturbation sweep (Table~\ref{tab:name_prompt_noise}), we observe a clear pattern: as text noise increases from $0\%$ to $60\%$, positive-region segmentation and accuracy decline while negative-region metrics (↓ better) improve, leaving the global score only mildly affected. Most of the damage occurs early at $20{\sim}40\%$ noise (p-mIoU $65.0$→$57.2$→$51.2$; p-cIoU $71.8$→$64.4$→$57.9$), and positive cumulative IoU then largely plateaus from $40\%$ to $60\%$ (p-cIoU $57.9$→$57.4$). 
This asymmetry indicates that mismatched prompts drive the model to back off to conservative masks, reducing spurious activations on background (n-mIoU $19.7$→$10.0$; n-cIoU $34.1$→$12.7$) while hurting recall on active regions, consistent with a fallback to visual cues when language becomes unreliable. 
Consequently, gIoU changes only slightly ($70.7$→$67.1$) and can obscure failure modes, making p/n-partitioned metrics the more informative diagnostics.

\subsection{Effects of Mask Annotation Noise}
With $k{\times}k$ morphological dilation applied to ground-truth masks (Table~\ref{tab:mask_edge_noise}), performance degrades chiefly on positive-region metrics and the aggregate score: p-mIoU drops steeply at moderate noise ($65.0$→$55.7$ at $9{\times}9$) and then continues to fall with diminishing increments ($49.1$ at $15{\times}15$, $44.8$ at $21{\times}21$), while gIoU declines from $70.7$→$68.2$→$63.7$→$60.3$. 
Negative-region metrics (↓ better) improve at moderate noise but show a non-monotone rebound at the highest level: n-mIoU $19.7$→$13.7$→$13.3$→$13.9$ and n-cIoU $34.1$→$20.3$→$19.7$→$29.9$. This pattern indicates that boundary dilation first discourages spurious background activations, pushing the model toward conservative masks with fewer false positives. 
Yet, once the noisy boundary band becomes too wide, supervision near edges turns conflicting, inducing boundary drift and shape jitter that hurt positive overlap and increase large area background overlap (hence the n-cIoU rebound). 
Acc changes little ($81.7$→$79.2$), suggesting the classifier is relatively insensitive compared with segmentation overlap metrics. At matched severities, mask noise is also more damaging than text noise (\textit{e.g.}, gIoU $60.3$ at $21{\times}21$ \textit{vs.}\ $67.1$ at Name-60; cf.\ Table~\ref{tab:name_prompt_noise}), underscoring the need for boundary-focused robustness and for reporting p/n-partitioned metrics over a single aggregate score.

\input{table/name60_mask21_noise}

\subsection{Evaluation of Robustness Strategies under Combined Noise Conditions}
Under the mixed setting (Name-60 + Mask-$21{\times}21$; Table~\ref{tab:name60_mask21_noise}), overall accuracy and positive-region IoUs drop across the board, and the ranking shifts relative to the single-noise sweeps (Table~\ref{tab:noise_all_single}): methods that filter sample-level noise (\textit{e.g.}, Co-teaching~\cite{han2018co}) no longer dominate, whereas losses that damp per-pixel label noise (GCE~\cite{zhang2018generalized}, SCE~\cite{wang2019symmetric}, APL~\cite{ye2023active}) and consistency-based training (NPN~\cite{sheng2024adaptive}) tend to rank higher on gIoU and the p-/n-partitioned IoUs. 
This aligns with the structure of multimodal corruption: name mismatches introduce instance-level grounding errors, whereas dilated masks introduce dense boundary bias; sample selection helps when labels are globally wrong but is brittle when many edge pixels are systematically corrupted. 
We also observe the familiar positive–negative trade-off; recovering active-region IoUs often increases background errors, so gIoU alone can misorder methods, making the p-/n-partitioned scores the more informative diagnostics. 

\textit{Observation 1} is the non-transitive robustness across noise types. Methods that prevail under text-only noise (\textit{e.g.}, Co-teaching at Name-40/60 in Table~\ref{tab:noise_all_single}) do not necessarily remain superior under mixed noise (Table~\ref{tab:name60_mask21_noise}). 
This is consistent with the premise of Co-teaching~\cite{han2018co}, which mitigates instance-level label contradictions but offers limited protection against dense, pixel-localized boundary corruption.

\textit{Observation 2} is a gradient-conflict effect under mixed noise. Prompt mis-grounding encourages conservative masks (shrinking foreground to avoid false activations), whereas boundary dilation encourages expansion near edges; pixel-wise robust losses (GCE~\cite{zhang2018generalized}, SCE~\cite{wang2019symmetric}, APL~\cite{ye2023active}) and dual-view consistency (NPN~\cite{sheng2024adaptive}) better reconcile these opposing update directions than sample filtering alone.
This interpretation matches the design goals of GCE/SCE/APL and NPN, now evidenced in a multimodal, pixel-level setting.

\textit{Observation 3} concerns metric sensitivity and benchmark coverage. Since positive and negative regions shift oppositely under mixed noise, aggregate gIoU can mask failures; reporting p-/n-partitioned mIoU/cIoU surfaces boundary-specific errors and background precision. 
The consistent re-ranking between Table~\ref{tab:noise_all_single} and Table~\ref{tab:name60_mask21_noise} validates the need for our multimodal protocols and partitioned metrics to fully characterize robustness in action-based video object segmentation.

\subsection{PMHM under Boundary and Mixed Noise}
As \textit{PMHM} is designed to address corrupted mask labels, we evaluated its performance under boundary and mixed noise conditions. In the boundary-only noise setting (kernel sizes $9$, $15$, and $21$), \textit{PMHM} consistently reduces background errors and improves global scores over the baseline. At the highest noise level ($21{\times}21$), it achieves a gIoU of $61.7$, outperforming the baseline's $60.3$ (Table~\ref{tab:mask_edge_noise}). 
In the mixed-noise setting (Name-60 + Mask-21), where severe text noise is introduced, \textit{PMHM's} performance (gIoU $59.9$) does not surpass the baseline FLac (gIoU $60.5$), as shown in Table~\ref{tab:name60_mask21_noise}. 
This confirms that while \textit{PMHM} resists boundary corruption, its gains fade when combined with significant text noise.

\begin{figure}[tb]
\small 
  \centering
  \includegraphics[width=\linewidth]{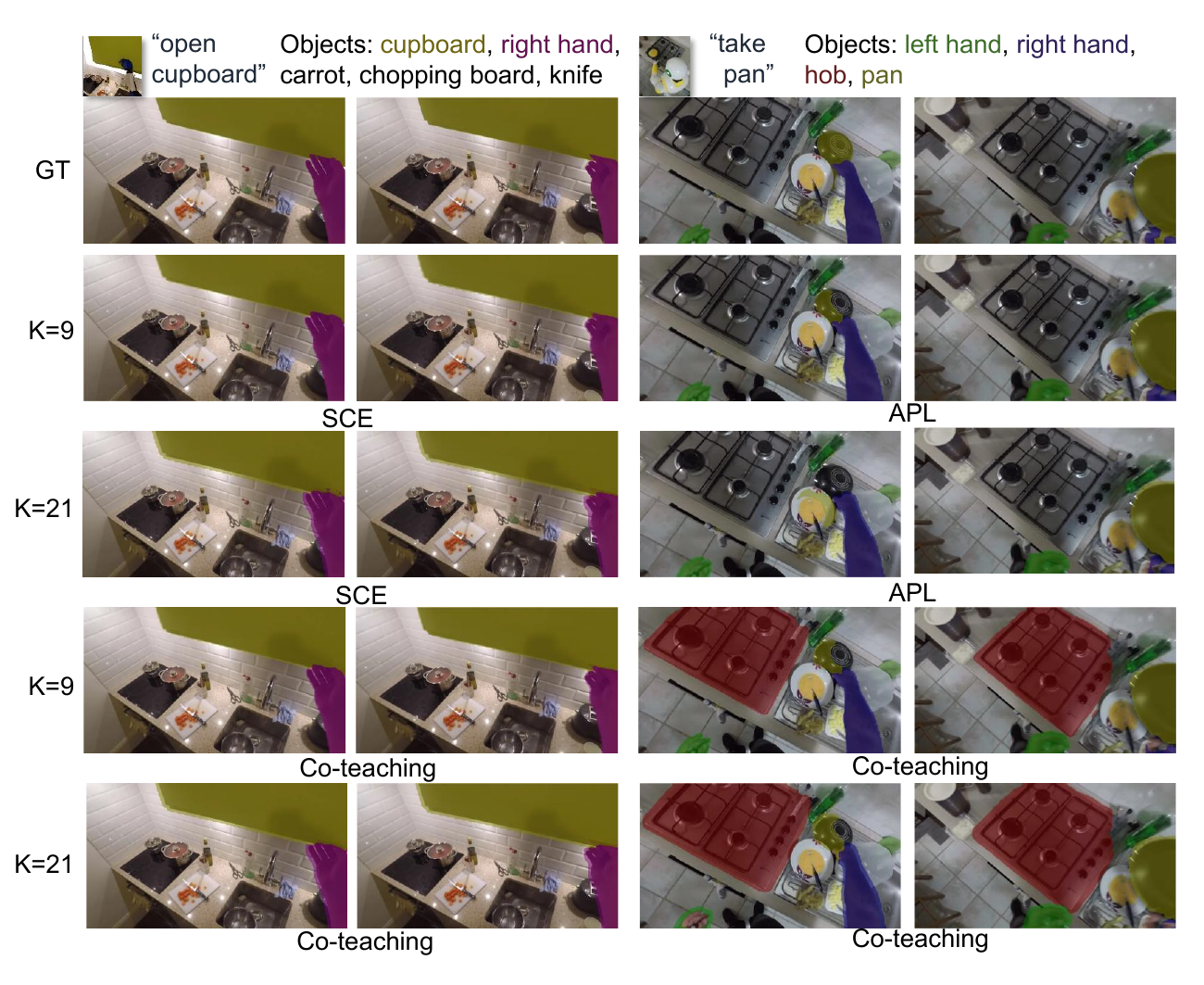}
  \caption{Mask annotation noise qualitative results on \textit{ActiSeg\mbox{-}NL}. Larger kernels thicken boundaries, coarsen edges, and introduce redundant regions, revealing trade-offs between foreground coverage and background precision.}
  \label{fig:mask_edge_qualitative}
  \vspace{-4mm}
\end{figure}

\subsection{Implications for Embodied Perception}
For tasks that require strong background suppression, the more balanced profiles of GCE and SCE are preferable. They reduce background leakage while maintaining active-region coverage, consistent with the re-ranking between Table~\ref{tab:noise_all_single} and Table~\ref{tab:name60_mask21_noise}. When language prompts are unreliable, Co-teaching with extended training preserves foreground under text noise, yet it provides limited protection against dense boundary corruption. It should therefore be paired with boundary-aware checks before execution. 
APL shows cleaner object edges at small kernel sizes but may overextend coverage at large kernel sizes, which warrants caution in tight workspaces. ELR increases positive-region IoUs and can support robust pick-up under occlusion, although the reduction in background precision calls for uncertainty gating. In practice, the partitioned metrics are task-aligned proxies: p-cIoU and p-mIoU reflect contact placement accuracy, and n-cIoU and n-mIoU reflect collision risk. Their joint trend provides a more reliable go/no-go signal than a single aggregate score. The qualitative patterns in Fig.~\ref{fig:mask_edge_qualitative} translate into concrete risks and design choices for robotic manipulation and closed-loop interaction. Boundary thickening and edge coarsening can cause grasp-pose drift and increase collision risk near contact surfaces. Spurious activation on inactive objects, as seen on the hob in the ``take pan'' sequence, can mislead target selection.

%% file: table/noise_all_single.tex
\begin{table*}[t]
\centering
\caption{Single-noise benchmark for action-based video object segmentation.
Each subtable fixes the noise type and level/kernel:
Name-20/40/60 (clean masks); Mask-$k\!\in\!\{9,15,21\}$ (clean names). PMHM targets mask noise and is therefore evaluated only under noisy mask conditions in this table.}

\vspace{-5pt}
\label{tab:noise_all_single}

\begin{threeparttable}
\scriptsize
\setlength{\tabcolsep}{2pt}
\renewcommand{\arraystretch}{0.92}

\begin{subtable}[t]{0.49\textwidth}
  \centering
  \captionsetup{font={scriptsize}}
  \caption{Name-20 (mask clean)}
  \vspace{-4pt}
  \resizebox{\textwidth}{!}{%
  \begin{tabular}{l|cccccc}
  \toprule
  Method & p-mIoU$\uparrow$ & n-mIoU$\downarrow$ & p-cIoU$\uparrow$ & n-cIoU$\downarrow$ & gIoU$\uparrow$ & Acc$\uparrow$ \\
  \midrule
  Co-teaching\cite{han2018co}    & 63.2 & 18.5 & 71.1 & 35.2 & 70.3 & 80.9 \\
  SCE\cite{wang2019symmetric}    & 63.8 & 25.3 & 67.3 & 36.9 & 50.4 & 67.4 \\
  GCE\cite{zhang2018generalized} & 59.1 & 24.1 & 66.8 & 37.2 & 46.9 & 66.3 \\
  APL\cite{ye2023active}         & 62.4 & 25.7 & 67.0 & 39.0 & 49.1 & 66.8 \\
  ELR\cite{liu2020early}          & 56.2 & 12.2 & 60.6 & 12.3 & 69.3 & 79.0\\
  NPN\cite{sheng2024adaptive}      & 56.1 & 12.4 & 60.5 & 13.8 & 69.1 & 78.7 \\
  \bottomrule
  \end{tabular}}%
\end{subtable}\hfill
\begin{subtable}[t]{0.49\textwidth}
  \centering
  \captionsetup{font={scriptsize}}
  \caption{Name-40 (mask clean)}
  \vspace{-4pt}
  \resizebox{\textwidth}{!}{%
  \begin{tabular}{l|cccccc}
  \toprule
  Method & p-mIoU$\uparrow$ & n-mIoU$\downarrow$ & p-cIoU$\uparrow$ & n-cIoU$\downarrow$ & gIoU$\uparrow$ & Acc$\uparrow$ \\
  \midrule
  Co-teaching\cite{han2018co}    & 63.0 & 17.4 & 70.7 & 29.1 & 70.6 & 81.2 \\
  SCE\cite{wang2019symmetric}    & 63.7 & 25.7 & 67.7 & 36.9 & 47.5 & 64.7 \\
  GCE\cite{zhang2018generalized} & 61.3 & 25.1 & 65.4 & 36.9 & 44.9 & 63.6 \\
  APL\cite{ye2023active}         & 62.7 & 23.8 & 65.6 & 35.4 & 47.2 & 65.3 \\
  ELR\cite{liu2020early}         & 54.7 & 10.1 & 59.0 & 7.4 & 69.5 & 78.9 \\
  NPN\cite{sheng2024adaptive}      & 53.6 & 10.4 & 56.9 & 8.8 & 68.8 & 76.9 \\
  \bottomrule
  \end{tabular}}%
\end{subtable}
\vspace{2pt}

\begin{subtable}[t]{0.49\textwidth}
  \centering
  \captionsetup{font={scriptsize}}
  \caption{Name-60 (mask clean)}
  \vspace{-4pt}
  \resizebox{\textwidth}{!}{%
  \begin{tabular}{l|cccccc}
  \toprule
  Method & p-mIoU$\uparrow$ & n-mIoU$\downarrow$ & p-cIoU$\uparrow$ & n-cIoU$\downarrow$ & gIoU$\uparrow$ & Acc$\uparrow$ \\
  \midrule
  Co-teaching\cite{han2018co}    & 60.2 & 13.8 & 68.8 & 23.0 & 71.2 & 80.9 \\
  SCE\cite{wang2019symmetric}    & 58.3 & 12.8 & 67.1 & 18.7 & 70.7 & 80.1 \\
  GCE\cite{zhang2018generalized} & 56.7 & 11.5 & 65.3 & 15.0 & 70.0 & 81.1 \\
  APL\cite{ye2023active}         & 57.8 & 11.6 & 66.3 & 15.8 & 70.9 & 80.3 \\
  ELR\cite{liu2020early}         & 48.5 & 9.8 & 51.8 & 11.8 & 66.7 & 72.2 \\
  NPN\cite{sheng2024adaptive}      & 47.6 & 9.2 & 51.6 & 7.9 & 66.3 & 71.8 \\
  \bottomrule
  \end{tabular}}%
\end{subtable}\hfill
\begin{subtable}[t]{0.49\textwidth}
  \centering
  \captionsetup{font={scriptsize}}
  \caption{Mask-$k{=}9$ (name clean)}
  \vspace{-4pt}
  \resizebox{\textwidth}{!}{%
  \begin{tabular}{l|cccccc}
  \toprule
  Method & p-mIoU$\uparrow$ & n-mIoU$\downarrow$ & p-cIoU$\uparrow$ & n-cIoU$\downarrow$ & gIoU$\uparrow$ & Acc$\uparrow$ \\
  \midrule
  Co-teaching\cite{han2018co}    & 60.6 & 23.4 & 69.8 & 51.1 & 65.1 & 80.1 \\
  SCE\cite{wang2019symmetric}    & 58.7 & 18.5 & 67.4 & 33.1 & 67.0 & 81.7 \\
  GCE\cite{zhang2018generalized} & 58.8 & 18.5 & 67.5 & 34.2 & 67.1 & 81.6 \\
  APL\cite{ye2023active}         & 58.8 & 18.7 & 68.4 & 33.9 & 67.0 & 81.3 \\
  ELR\cite{liu2020early}         & 57.1 & 18.8 & 66.0 & 32.3 & 64.6 & 78.0 \\
  NPN\cite{sheng2024adaptive}      & 58.8 & 20.6 & 67.5 & 37.1 & 65.4 & 78.6 \\
  \cellcolor[gray]{.9}{PMHM}      & \cellcolor[gray]{.9}{55.6} & \cellcolor[gray]{.9}{13.6} & \cellcolor[gray]{.9}{64.3} & \cellcolor[gray]{.9}{21.5} & \cellcolor[gray]{.9}{67.8}  & \cellcolor[gray]{.9}{81.1} \\
  \bottomrule
  \end{tabular}}%
\end{subtable}

\vspace{2pt}

\begin{subtable}[t]{0.49\textwidth}
  \centering
  \captionsetup{font={scriptsize}}
  \caption{Mask-$k{=}15$ (name clean)}
  \vspace{-4pt}
  \resizebox{\textwidth}{!}{%
  \begin{tabular}{l|cccccc}
  \toprule
  Method & p-mIoU$\uparrow$ & n-mIoU$\downarrow$ & p-cIoU$\uparrow$ & n-cIoU$\downarrow$ & gIoU$\uparrow$ & Acc$\uparrow$ \\
  \midrule
  Co-teaching\cite{han2018co}    & 53.4 & 21.3 & 65.4 & 47.0 & 60.9 & 79.5 \\
  SCE\cite{wang2019symmetric}    & 51.4 & 14.7 & 63.4 & 28.4 & 64.2 & 82.4 \\
  GCE\cite{zhang2018generalized} & 51.5 & 13.9 & 62.1 & 28.4 & 64.6 & 82.1 \\
  APL\cite{ye2023active}         & 51.6 & 13.6 & 61.5 & 24.0 & 64.7 & 82.5 \\
  ELR\cite{liu2020early}         & 51.2 & 17.4 & 62.0 & 30.2 & 62.0 & 79.1 \\
  NPN\cite{sheng2024adaptive}      & 52.2 & 18.2 & 62.2 & 34.9 & 61.9 & 78.6 \\
  \cellcolor[gray]{.9}{PMHM}      & \cellcolor[gray]{.9}{48.7} & \cellcolor[gray]{.9}{11.8} & \cellcolor[gray]{.9}{59.5} & \cellcolor[gray]{.9}{19.2} & \cellcolor[gray]{.9}{64.6}  & \cellcolor[gray]{.9}{81.2} \\
  \bottomrule
  \end{tabular}}%
\end{subtable}\hfill
\begin{subtable}[t]{0.49\textwidth}
  \centering
  \captionsetup{font={scriptsize}}
  \caption{Mask-$k{=}21$ (name clean)}
  \vspace{-4pt}
  \resizebox{\textwidth}{!}{%
  \begin{tabular}{l|cccccc}
  \toprule
  Method & p-mIoU$\uparrow$ & n-mIoU$\downarrow$ & p-cIoU$\uparrow$ & n-cIoU$\downarrow$ & gIoU$\uparrow$ & Acc$\uparrow$ \\
  \midrule
  Co-teaching\cite{han2018co}    & 46.7 & 18.7 & 59.5 & 45.2 & 58.1 & 79.5 \\
  SCE\cite{wang2019symmetric}    & 44.1 & 12.3 & 55.8 & 24.8 & 60.6 & 80.0 \\
  GCE\cite{zhang2018generalized} & 44.7 & 12.5 & 57.6 & 22.1 & 60.9 & 79.9 \\
  APL\cite{ye2023active}         & 45.7 & 14.3 & 58.4 & 30.3 & 60.8 & 82.0 \\
  ELR\cite{liu2020early}          & 45.2 & 16.3 & 57.4 & 34.2 & 58.4 & 78.0\\
  NPN\cite{sheng2024adaptive}      & 46.2 & 17.2 & 58.5 & 37.0 & 58.9 & 78.0 \\
  \cellcolor[gray]{.9}{PMHM}      & \cellcolor[gray]{.9}{43.8} & \cellcolor[gray]{.9}{11.0} & \cellcolor[gray]{.9}{56.6} & \cellcolor[gray]{.9}{20.9} & \cellcolor[gray]{.9}{61.7}  & \cellcolor[gray]{.9}{80.1} \\
  \bottomrule
  \end{tabular}}%
\end{subtable}

\end{threeparttable}
\vspace{-14pt}
\end{table*}

%% file: table/name60_mask21_noise.tex
\begin{table}[t]
\centering
\setlength{\tabcolsep}{3pt}  
\caption{Name noise $60\%$ + Mask kernel $21{\times}21$ on action-based video object segmentation (mixed setting).
$\uparrow$/$\downarrow$ denotes that higher/lower is better.}
\vspace{-2pt}
\label{tab:name60_mask21_noise}
\begin{tabular}{l|cccccc}
\toprule[1pt]
Method  & p-mIoU$\uparrow$ & n-mIoU$\downarrow$ & p-cIoU$\uparrow$ & n-cIoU$\downarrow$ & gIoU$\uparrow$ & Acc$\uparrow$ \\
\midrule[1pt]
FLac\cite{ouyang2024actionvos}      & 37.8 & 7.6 & 49.7 & 11.2 & 60.5 & 76.6 \\
Co-teaching\cite{han2018co}    & 36.2 & 10.9 & 48.7 & 24.3 & 57.9 & 70.4 \\
SCE\cite{wang2019symmetric}         & 36.4 & 6.5 & 47.8 & 6.0 & 60.9 & 72.9 \\
GCE\cite{zhang2018generalized}      & 38.7 & 7.9 & 49.7 & 11.8 & 61.1 & 75.4 \\
APL\cite{ye2023active}              & 36.5 & 7.1 & 48.1 & 9.9 & 60.5 & 73.6 \\
ELR\cite{liu2020early}              & 41.5 & 15.9 & 51.2 & 33.3 & 48.8 & 72.4 \\
NPN\cite{sheng2024adaptive}            & 37.6 & 8.0 & 49.7 & 14.0 & 60.2 & 75.2 \\
\cellcolor[gray]{.9}{PMHM} &
\cellcolor[gray]{.9}{36.8} &
\cellcolor[gray]{.9}{8.4} &
\cellcolor[gray]{.9}{47.1} &
\cellcolor[gray]{.9}{14.6} &
\cellcolor[gray]{.9}{59.9}  &
\cellcolor[gray]{.9}{74.1} \\
\bottomrule[1pt]
\end{tabular}

\vspace{-14pt}
\end{table}

%% file: Contents/Conclusion.tex
We introduce \textit{ActiSeg-NL}, a benchmark to assess action-based video object segmentation under text and mask noise, towards advancing embodied intelligence. Our analysis shows that model performance degrades significantly with noise, especially from masks. 
We find that robust learning strategies present clear trade-offs: some preserve foregrounds under text noise (Co-teaching), others improve overlap (APL), or achieve a superior balance in mixed conditions (GCE, SCE). 
Future work seeks to solve this trade-off by integrating vision language foundation models to drive boundary-focused objectives and uncertainty-guided consistency under real-world noise, together with systematic validation in closed-loop robotic systems.
%